\documentclass[letterpaper]{article}
\usepackage[preprint]{paperstyle}
\usepackage[hyphens]{url}
\usepackage{graphicx}
\usepackage{natbib}
\usepackage{caption}
\usepackage{amsmath,amssymb}
\usepackage{booktabs}
\usepackage[table]{xcolor}
\usepackage{multirow}

\title{CARVE: Cross-Slice Anisotropic Reallocation of Visual Evidence\\for Efficient 3D Medical Volume Understanding}
\author{
    Zhenyu Yi\textsuperscript{\rm 1},
    Qiang Hu\textsuperscript{\rm 2},
    Zhenhao Li\textsuperscript{\rm 1},
    Jiaxuan Zhao\textsuperscript{\rm 1},
    Yusong Sun\textsuperscript{\rm 1},
    Lichi Zhang\textsuperscript{\rm 1}\thanks{Corresponding author.}
}
\affiliations{
    \textsuperscript{\rm 1}School of Biomedical Engineering, Shanghai Jiao Tong University\\
    \textsuperscript{\rm 2}Wuhan National Laboratory for Optoelectronics, Huazhong University of Science and Technology
}
 
\begin{document}
\maketitle

% ============================================================ 
\begin{abstract}
% Slice-based multimodal large language models (MLLMs) process a 3D medical volume by encoding each 2D slice and feeding the entire slice sequence into the language model. This design inherits mature 2D encoders, but the thousands of visual tokens it produces burden the LLM backbone, while much of that budget re-encodes the same structure on adjacent slices. Our scaling analyses on two 3D medical VQA benchmarks show diminishing returns: cost keeps rising while accuracy saturates, and at comparable token counts, increasing in-plane resolution is more effective than adding slices. 
% However, this slice-wise formulation produces thousands of visual tokens that burden the LLM backbone, and repeatedly capturing overlapping medical visual evidence across adjacent slices.
Slice-based MLLMs leverage mature 2D encoders by representing 3D volumes as sequences of 2D slices. 
However, this slice-wise formulation produces thousands of visual tokens that burden the LLM backbone, many of which capture overlapping visual evidence across adjacent slices.
To understand how effectively a growing visual token budget improves performance, we perform scaling analyses on two 3D medical VQA benchmarks and find diminishing returns: cost keeps rising while accuracy saturates, and improving in-plane resolution is more effective than adding slices at comparable budgets.
The budget should therefore be allocated more selectively rather than simply enlarged, yet most token compression methods are designed for 2D images or videos, where redundancy arises from spatial layout or temporal motion rather than from near-duplicate content along the depth axis.
We present CARVE, a training-free framework that compresses visual tokens prior to LLM inference and casts token reduction as budget-constrained 2.5D allocation. CARVE partitions the depth axis into coherent windows and allocates tokens non-uniformly according to normalized cross-slice evidence. 
Under a shared budget, CARVE builds spatial anchors on representative slices and retrieves locally varying evidence from the full volume, then merges remaining eligible tokens into nearby anchors within each window.
Removing roughly 80\% of the visual tokens on Hulu-Med-7B, CARVE leads all compression baselines on every AMOS-MM report-generation metric, with 6.2 points higher retention of full-token quality than the strongest baseline, and preserves 98.1\% of full-token performance across three VQA benchmarks.
\end{abstract}

% ============================================================
\section{Introduction}
\label{sec:intro}

Medical multimodal large language models (MLLMs) are increasingly applied to whole 3D volumes. Rather than learning native volumetric representations~\cite{wu2024towards,bai2024m3d}, most recent open-source systems reuse a mature 2D vision stack over ordered axial images~\cite{jiang2025hulu,jiang2025omniv,sellergren2025medgemma}: each slice is encoded independently, the patch features are concatenated, and cross-slice modeling is deferred to the projector and LLM, optionally with 3D-aware position indices~\cite{wang2024qwen2}. We target these slice-based models, which inherit strong 2D pretraining and support variable slice counts but realize a volume primarily by scaling the visual sequence: a moderate $32\!\times\!512\!\times\!512$ input already produces ${\sim}$10{,}367 visual tokens on Hulu-Med-7B, while cross-slice relations remain unresolved inside the 2D encoder. Along the depth axis, neighboring CT slices share broad anatomy, yet small structures, boundaries, or abnormalities may change over only a few slices. The sequence reaching the expensive LLM is therefore both long and structurally redundant.

What must survive compression also depends on the task. A VQA query may target coarse anatomy, a localized attribute, or a cross-slice relation, while 3D report generation must summarize spatially distributed findings rather than only one prompted region~\cite{hamamci2024ct2rep,li2025towards,yu2025medframeqa}. Compression must consequently maintain a spatial scaffold for recurring context while preserving a recall path for localized evidence anywhere in the volume.
\begin{figure}[t]
\centering  
\includegraphics[width=\columnwidth]{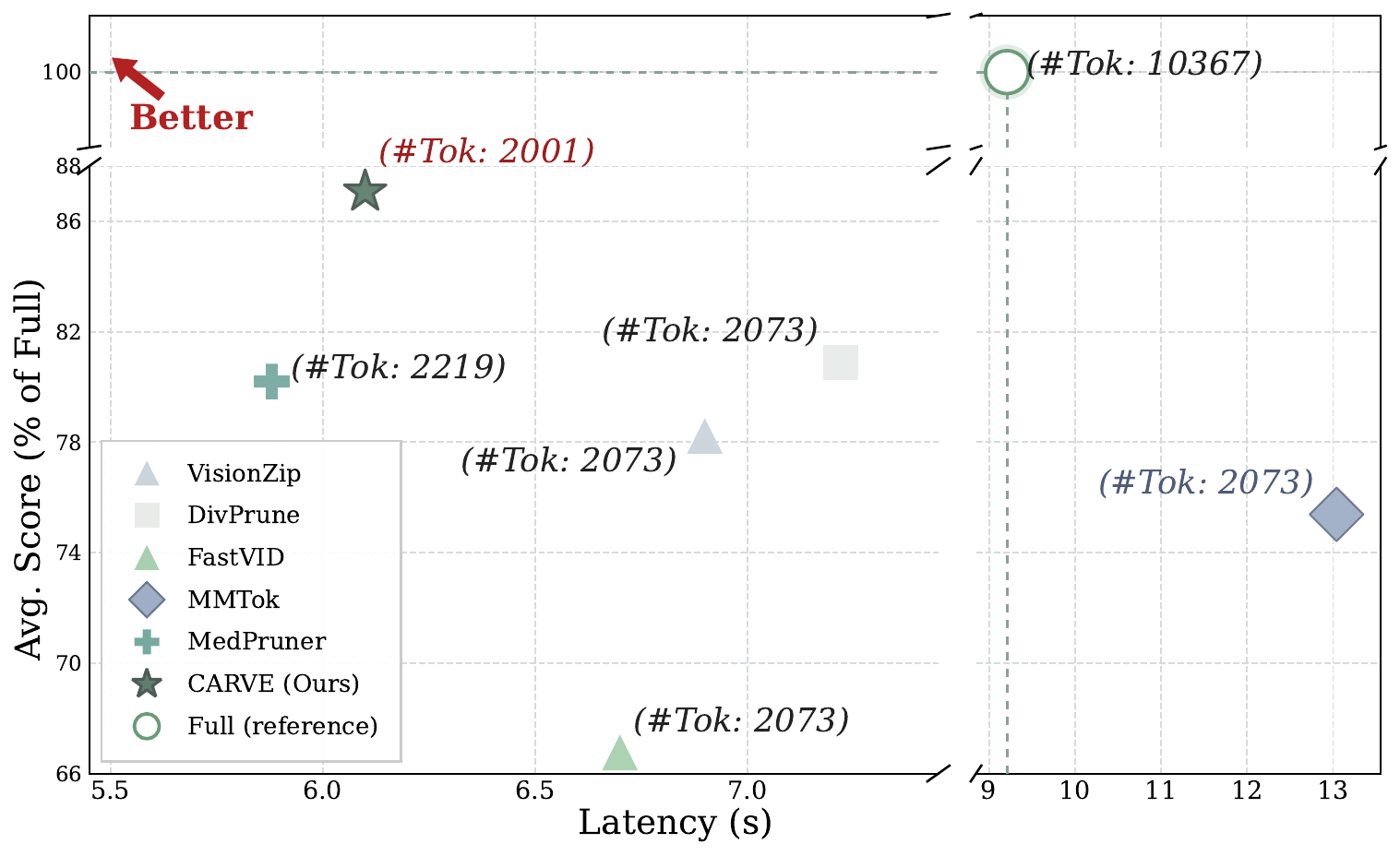} 
\caption{Quality--latency trade-off on AMOS-MM report generation with Hulu-Med-7B at a 20\% target keep ratio. Labels give the retained visual token count (\#Tok) per volume; quality averages relative retention over the four report metrics of Table~\ref{tab:amos_report_main}. CARVE establishes the strongest quality--efficiency frontier among compressed methods.}
\label{fig:tradeoff} 
\end{figure} 
\begin{figure*}[t]
\centering 
\includegraphics[width=0.98\textwidth]{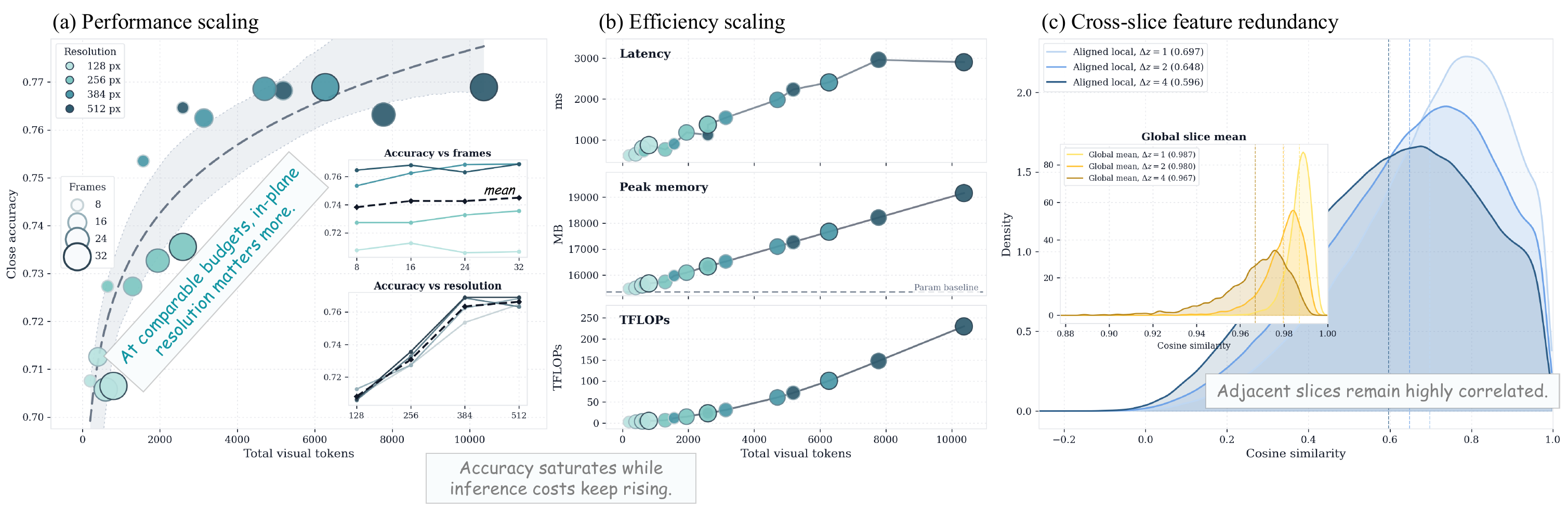}
\caption{Scaling analysis of slice-based 3D medical VLMs on AMOS-MM VQA. (a) Accuracy across slice-resolution combinations, with bubble size denoting slice count; in-plane resolution helps more than added slices at comparable budgets. (b) Accuracy against latency, peak memory, and estimated TFLOPs. (c) Adjacent-slice feature correlation at aligned local positions and in global slice means.}
\label{fig:intro}
\end{figure*} 
Those requirements concern how a visual budget is spent, not only how large it is. We first ask: \emph{Do additional 3D visual tokens contribute equally, wherever they are allocated?} Figure~\ref{fig:intro} shows they do not: on AMOS-MM, accuracy saturates while latency, memory, and FLOPs keep rising, in-plane resolution beats extra slices at matched token counts, and adjacent slices stay highly similar in feature space. The bottleneck is thus not merely an oversized sequence, but a budget distributed without regard to the depth/in-plane asymmetry of the volume.

Token compression has so far been developed for other input types. Image methods rank or merge tokens over a flat 2D layout~\cite{yang2025visionzip,dhouib2025pact,deng2026scope,zhang2026beyond,dong2025mmtok}, video methods exploit temporal redundancy~\cite{shen2026fastvid,huang2025prunevid}, and geometry-aware methods operate on point or voxel representations~\cite{li2026segpruner,li2026geometry,huang2025zero}. None of these inputs presents an anisotropic depth axis that repeatedly re-images the same anatomy, so their objectives were never required to divide one budget between two axes of unequal value, and they transfer to volumes only partially. In the medical setting, MedPruner and other slice-selection methods~\cite{liu2026medpruner,lian2026data,chen2025adapting} do exploit volumetric redundancy, filtering slices before retaining tokens within the survivors; because they act at whole-slice granularity, however, evidence on a discarded plane cannot be recovered.

Guided by these observations, we cast 3D medical token compression as \emph{anisotropic 2.5D budget allocation} and present CARVE (\textbf{C}ross-slice \textbf{A}nisotropic \textbf{R}eallocation of \textbf{V}isual \textbf{E}vidence). CARVE partitions the depth axis into feature-continuous windows, not to discard their non-representative slices, but to define where budgets are allocated and tokens can be folded safely. Under one target budget, intra-slice selection constructs multi-granular spatial anchors on representative slices, while global inter-slice retrieval searches the complete volume for locally changing evidence missed by those anchors; eligible remaining tokens are then folded into nearby anchors within the same window, whereas retrieved tokens remain independent. CARVE is training-free and operates pre-LLM, requiring no modification to the underlying medical MLLM.

In summary, our contributions are threefold:
\begin{enumerate}
  \item \textbf{Empirical analysis of 3D token scaling.} Across AMOS-MM and M3D-VQA, in-plane resolution outperforms adding slices at matched token counts and neighboring slices stay highly redundant in feature space, motivating non-uniform 2.5D allocation rather than uniform scaling.
  \item \textbf{A training-free pre-LLM reallocation framework.} We propose CARVE, which couples anisotropic window allocation with intra-slice anchor selection and global inter-slice evidence retrieval, followed by window-local token folding before the LLM.
  \item \textbf{Strong quality-efficiency trade-offs.} CARVE leads all compression baselines on every AMOS-MM report-generation metric and preserves near-full aggregate VQA performance at approximately 19\% retained tokens, giving the best quality--efficiency frontier among compressed methods (Fig.~\ref{fig:tradeoff}).
\end{enumerate}

% ============================================================ 

\section{Related Work}
\label{sec:related}

\subsection{3D Medical Vision-Language Models}

Medical VLMs have progressed from 2D biomedical assistants such as LLaVA-Med~\cite{li2023llava} toward volume-capable systems following two representation routes. Native volumetric models, including RadFM, M3D, Merlin, and CT-RATE foundation models~\cite{wu2024towards,bai2024m3d,blankemeier2024merlin,hamamci2026generalist}, learn 3D representations directly. Slice-based systems instead reuse 2D vision backbones over ordered axial images~\cite{jiang2025hulu,jiang2025omniv,sellergren2025medgemma}; their features are assembled downstream by the projector and LLM, sometimes with multimodal positional indexing such as M-RoPE~\cite{wang2024qwen2}. This route inherits strong pretrained encoders and naturally supports variable slice sequences, but its token count grows along both depth and in-plane axes while cross-slice interaction is deferred beyond the encoder.

Volume-level tasks expose complementary evidence requirements. CT2Rep and 3D brain-CT MLLMs generate reports from complete scans~\cite{hamamci2024ct2rep,li2025towards}, while MedFrameQA evaluates reasoning across multiple medical images~\cite{yu2025medframeqa}. Dual-stream MIL and cross-view alignment further couple global predictions with localized evidence for 3D diagnosis~\cite{yi2026brain,li2026mrl}. These works motivate maintaining both broad anatomical support and sparse local findings, but address modeling, supervision, or evaluation rather than drop-in sequence reduction. Efficiency-oriented systems instead prune consecutive slices~\cite{jiang2025omniv} or build the volume model itself for lower-cost interpretation, either learning instruction-conditioned token budgets jointly with the architecture during training~\cite{fang2026photon} or redesigning the encoder, tokenizer, or projector~\cite{lee2026read,xin2025med3dvlm,hamamci2026better}. All of these change the model; CARVE addresses the complementary setting of compressing a frozen slice-based model after its encoder, without task-specific training.

\subsection{Visual Token Compression}

LLaVA-PruMerge and VisionZip retain salient tokens while merging redundancy~\cite{shang2024llava,yang2025visionzip}; PACT combines pruning with bounded clustering at an early LLM layer~\cite{dhouib2025pact}. Training-free ranking methods use focal or hierarchical attention~\cite{jiang2024fopru,liu2026hiprune}, feature diversity or graph structure~\cite{alvar2025divprune,jiang2025kind}, and saliency-coverage or multimodal-coverage objectives~\cite{deng2026scope,dong2025mmtok}. Closer to volumetric compression, video methods exploit redundancy across ordered inputs and allocate budgets across segments~\cite{shen2026fastvid,huang2025prunevid,hyun2025multi,ma2025mmg} and geometry-aware 3D methods use spatial sampling or voxel compression~\cite{li2026segpruner,li2026geometry,huang2025zero}. Yet a medical slice axis is neither ordinary time nor a generic point/voxel set: it repeatedly observes aligned anatomy with sparse embedded changes, so none of these objectives determines how one budget should be divided between the depth and in-plane axes.
MedPruner is the closest training-free medical baseline: it filters slices before adaptively retaining tokens within the selected planes~\cite{liu2026medpruner}; trained 2D-encoder pruning and systematic slice-selection studies share this granularity~\cite{lian2026data,chen2025adapting}. CARVE differs on exactly this point: rather than committing to a subset of planes, it reallocates tokens across the anisotropic volume and keeps every slice eligible for retrieval. Placement also matters: ToMe and EViT alter token flow inside the vision encoder~\cite{bolya2022token,liang2022not} and FastV prunes after visual tokens enter the LLM~\cite{chen2024image}, whereas CARVE operates training-free at the post-encoder, pre-projector interface, coupling depth-windowed allocation, intra-slice anchors, and global inter-slice retrieval before LLM computation.
 
\iffalse
\begin{table}[t]
\centering

\setlength{\tabcolsep}{2.5pt}
\footnotesize
\begin{tabular}{@{}lc c c c l@{}}
\toprule
Method & \rotatebox{60}{Attn} & \rotatebox{60}{Sim} & \rotatebox{60}{Query} & \rotatebox{60}{Spatial} & Selection Strategy \\
\midrule
\multicolumn{6}{@{}l}{\textit{2D Image}} \\
FoPru & \checkmark & & & & Top-K \\
HiPrune & \checkmark & & & & Hierarchical \\
VisionZip & \checkmark & \checkmark & & & Select+Merge \\
SCOPE & \checkmark & \checkmark & & & Coverage$\times$Sal. \\
DivPrune & & \checkmark & & & Max-Min Div. \\
GPrune & & \checkmark & & & Graph+Top-K \\
MMTok & & \checkmark & \checkmark & & Max Coverage \\
CDPruner & & \checkmark & \checkmark & & Cond. DPP \\
\midrule
\multicolumn{6}{@{}l}{\textit{3D Scene}} \\
DTC & & \checkmark & & \checkmark & Voxel Merge \\
SeGPruner & \checkmark & \checkmark & & \checkmark & Top-K + FPS \\
\midrule
\multicolumn{6}{@{}l}{\textit{3D Medical}} \\
MedPruner & \checkmark & \checkmark & & & Filter+Nucleus \\
\textbf{CARVE} & \checkmark & \checkmark & & \checkmark & \textbf{Dual-Branch+Merge} \\
\bottomrule
\end{tabular*}
\caption{Scoring signals and compression strategies of training-free token pruning methods. CARVE jointly models attention, feature similarity, and 2.5D spatial continuity within a dual-branch select-and-merge pipeline.}
\label{tab:method_comparison}
\end{table} 
\fi 

% ============================================================
\section{Method}  
\label{sec:method}

\begin{figure*}[t]
\centering      
\includegraphics[width=0.9\textwidth]{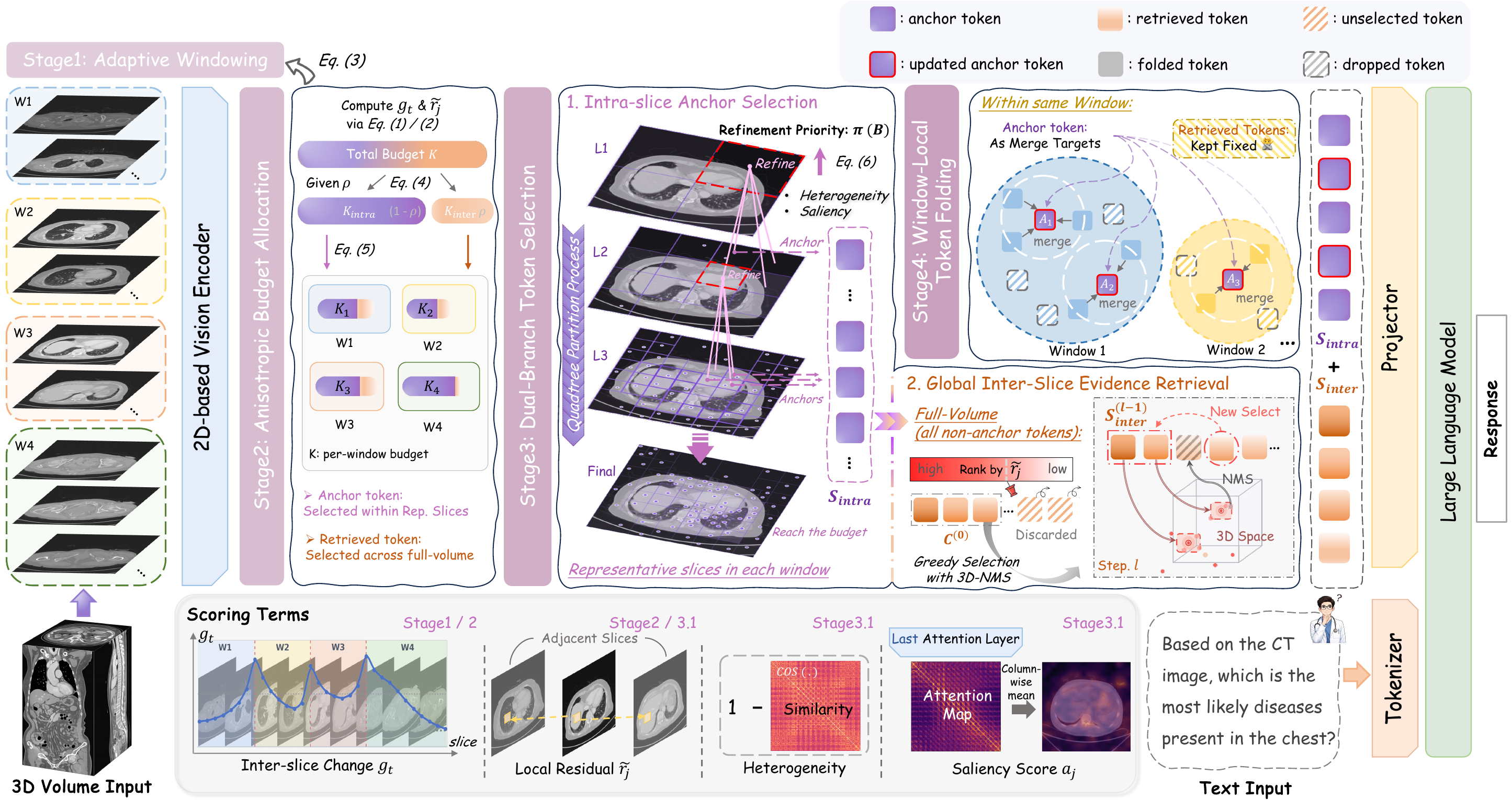}
\caption{Overview of CARVE. Stage 1 partitions slice features into adaptive depth windows. Stage 2 allocates the budget anisotropically across windows and between the two selection roles. Stage 3 couples intra-slice anchor selection on representative slices with global inter-slice retrieval of dispersed deviations from the full volume. Stage 4 folds eligible remaining tokens into same-window anchors; retained tokens keep their positions before the frozen projector and LLM.}
\label{fig:method}
\end{figure*}

We present \textbf{CARVE}, a training-free token compression framework that performs anisotropical budget allocation for slice-based 3D medical MLLMs. Repeated structures across adjacent slices can be represented economically by anchors on a few representative slices, whereas localized deviations may appear anywhere along the depth axis and require evidence retrieval from the full volume. A single flat ranking cannot fill both roles at once, since it may repeatedly select correlated tokens while sacrificing spatial coverage of discarded regions. CARVE therefore divides the target budget between \emph{intra-slice anchor selection} and \emph{global inter-slice evidence retrieval}, using adaptive depth windows to coordinate allocation and local token folding between them (Fig.~\ref{fig:method}).
% We present \textbf{CARVE}, a training-free token compression framework for slice-based 3D medical MLLMs. Repeated cross-slice structure can be represented economically by anchors on a few representative slices, whereas localized deviations may occur anywhere along the depth axis and are reachable only by a path spanning the full volume; a single flat ranking cannot fill both roles at once, since it can repeatedly select correlated tokens while weakening spatial support for the discarded ones. CARVE therefore divides one target budget between \emph{intra-slice anchor selection} and \emph{global inter-slice evidence retrieval}, using adaptive depth windows to coordinate allocation and local folding between them (Fig.~\ref{fig:method}).

\subsection{Problem Formulation}
\label{sec:formulation}

Consider a 3D medical volume with $T$ slices. A 2D vision encoder produces $N$ features $\mathcal{F}=\{\mathbf{f}_j\}_{j=1}^{N}$ with coordinates $\mathbf{c}_j=(z_j,u_j,v_j)$. A frozen multimodal projector $P$ would ordinarily map them to LLM visual tokens $\mathbf{x}_j=P(\mathbf{f}_j)$. Inserted between the encoder and $P$, CARVE compresses $\mathcal{F}$ under a target budget $K=\lfloor rN\rfloor$ for target keep ratio $r\in(0,1]$. It constructs complementary sets $\mathcal{S}_{\mathrm{intra}}$ of spatial anchors and $\mathcal{S}_{\mathrm{inter}}$ of independently kept cross-slice evidence; we call their members \emph{anchor tokens} and \emph{retrieved tokens}.

\subsection{Cross-Slice Profiling and Adaptive Windowing} 
\label{sec:window}

CARVE profiles depth-wise feature change at two scales: a slice-level signal for coherent depth neighborhoods, and a token-level residual for local deviations within them.

\paragraph{Inter-slice change.}
For each slice $t$, we mean-pool and $\ell_2$-normalize its raw encoder features into $\bar{\mathbf{f}}_t \in \mathbb{R}^d$, and define
\begin{equation}
    g_t = 1 - \cos(\bar{\mathbf{f}}_t,\; \bar{\mathbf{f}}_{t+1}), \quad t = 1, \ldots, T{-}1.
    \label{eq:inter_dist}
\end{equation}
Large $g_t$ marks a feature discontinuity along the slice axis; low-change runs indicate where broad structure is repeatedly encoded across neighboring slices.

\paragraph{Slice-normalized local residual.}
To separate slice-wide change from localized residuals, we compare each token with aligned neighbors and subtract its slice median:
\begin{equation}
    \begin{aligned}
    r_j &= 1-\frac{1}{|\mathcal{N}_z(j)|}
    \sum_{j'\in\mathcal{N}_z(j)}\cos(\mathbf{f}_j,\mathbf{f}_{j'}),\\
    \widetilde r_j &= \max\!\left(0,\ r_j-
    \operatorname*{median}_{k\in\mathrm{slice}(j)} r_k\right),
    \end{aligned}
    \label{eq:slicenorm}
\end{equation}
where $\mathcal{N}_z(j)$ contains available tokens at the same in-plane coordinate on slices $z_j\!\pm\!1$. Median subtraction removes slice-wide shifts, leaving $\widetilde r_j$ as a local cross-slice deviation score for evidence retrieval.

\paragraph{Adaptive windows.}
CARVE places a candidate window boundary after slice $t$ when
\begin{equation}
    g_t > \mu_g + \tau\sigma_g,
    \qquad t=1,\ldots,T-1,
    \label{eq:window_boundary}
\end{equation}
where $\mu_g$ and $\sigma_g$ are the mean and standard deviation of the inter-slice change scores, and $\tau$ controls boundary sensitivity. These boundaries partition the volume into windows $\{\mathcal{W}_s\}_{s=1}^{S}$ that share an allocation and folding domain. Importantly, a window is not a slice-pruning unit: every slice remains eligible for global evidence retrieval.

\subsection{Anisotropic Budget Allocation}
\label{sec:budget}

The target budget must preserve a spatial scaffold without closing the recall path across depth. We therefore reserve a global inter-slice budget and assign the remainder to intra-slice anchors:
\begin{equation}
    K_{\mathrm{inter}}=\lfloor\rho K\rfloor,\qquad
    K_{\mathrm{intra}}=K-K_{\mathrm{inter}}.
    \label{eq:branch_budget}
\end{equation}
Here $\rho\in[0,1]$ is the fraction reserved for cross-slice recall, and $K_{\mathrm{inter}}$ is an upper budget for $\mathcal{S}_{\mathrm{inter}}$.

We then distribute $K_{\mathrm{intra}}$ non-uniformly across windows. Windows with stronger depth variation, larger local residuals, or longer spans receive more anchors. To compare these signals on a common scale, let $\mathcal{N}(v_s)$ denote min--max normalization of statistic $v_s$ over the current volume; we use the fixed score
\begin{equation}
    \begin{aligned}
    w_s&=\mathcal{N}(\bar g_s)
    +\mathcal{N}(\bar r_s)
    +\tfrac{1}{2}\mathcal{N}(\log(1+n_s)),
    \\[-1pt]
    K_s&=\operatorname{LR}_s\!\left[
    K_{\mathrm{intra}}\,\mathrm{softmax}_s(w_s)\right],
    \end{aligned}
    \label{eq:window_allocation}
\end{equation}
Here $\bar g_s$ averages the change scores within $\mathcal{W}_s$, $\bar r_s$ averages $\widetilde r_j$ over its tokens, $n_s=|\mathcal{W}_s|$, and $\operatorname{LR}$ denotes largest-remainder integerization, so that $\sum_s K_s=K_{\mathrm{intra}}$.

\subsection{Intra-Slice Anchor Selection}
\label{sec:intra} 
 
The intra-slice path converts each window budget into a compact spatial scaffold. Within $\mathcal{W}_s$, it ranks slices by the mean cosine similarity of their pooled features to those of the other slices and distributes $K_s$ across the top-ranked representatives. Their number is $m_s=\mathrm{clamp}(\mathrm{round}(K_s/K_{\mathrm{slice}}),1,\min(n_s,K_s))$ for a target anchor density $K_{\mathrm{slice}}$, and $K_s$ is split as evenly as possible across them. This avoids spending anchors on many near-duplicate slices.

\paragraph{Multi-granular in-plane partition.}
On each representative slice, CARVE recursively refines the $H\times W$ token grid. Let $a_j$ be the attention saliency of token $j$, obtained by averaging the last-layer encoder self-attention $A^{(L,h)}_{qj}$ over all query positions $q$ and heads $h$; this reuses an existing attention artifact and needs no dedicated class token. Refinement favors blocks that are both internally heterogeneous and encoder-salient:
\begin{equation}
    \pi(\mathcal{B}) =
    \mathcal{N}\!\left(1-\frac{1}{|\mathcal{B}|}
    \sum_{j\in\mathcal{B}}\cos(\mathbf{f}_j,\bar{\mathbf{f}}_{\mathcal{B}})\right)
    +\mathcal{N}\!\left(\max_{j\in\mathcal{B}}a_j\right),
    \label{eq:priority}
\end{equation}
where $\bar{\mathbf{f}}_{\mathcal{B}}$ is the mean feature of block $\mathcal{B}$; the heterogeneity and attention terms are normalized independently over the current slice. At each step, the highest-priority block is split until the allocated budget is reached; the highest-attention token in each final block becomes its anchor. The resulting union $\mathcal{S}_{\mathrm{intra}}$ preserves multi-granular in-plane support. We realize this budgeted refinement with a standard best-first quadtree~\cite{hyun2025multi}.

\subsection{Global Inter-Slice Evidence Retrieval}
\label{sec:inter}
 
Representative-slice anchors efficiently summarize recurring structure, but by construction cannot guarantee that localized deviations on other slices survive. The inter-slice path therefore searches all non-anchor tokens using the residual in Eq.~\ref{eq:slicenorm}.

\paragraph{Hard residual ranking.}
Let $\mathcal{P}_{\mathrm{inter}}=\{1,\ldots,N\}\setminus\mathcal{S}_{\mathrm{intra}}$ be the full-volume retrieval pool and $s_j$ the volume-normalized score derived from $\widetilde r_j$. We first isolate the high-residual portion of this pool:
\begin{equation}
    \begin{aligned}
    M_{\mathrm{cand}}&=\max\!\left(
    K_{\mathrm{inter}},\left\lceil0.25|\mathcal{P}_{\mathrm{inter}}|\right\rceil\right),\\
    \mathcal{C}^{(0)}&=\left\{j\in\mathcal{P}_{\mathrm{inter}}:\
    \operatorname{rank}_{\downarrow}(s_j)
    \le M_{\mathrm{cand}}\right\}.
    \end{aligned}
    \label{eq:inter_candidates}
\end{equation}
This keeps the strongest quartile of deviations, and never fewer than $K_{\mathrm{inter}}$ candidates, for the subsequent spatial de-duplication.

\paragraph{Greedy selection with 3D-NMS.}
High residual scores can cluster around the same physical region. To prevent redundant retrieval, we define the spacing-aware suppression neighborhood of candidate $i$ as
\begin{equation}
    \mathcal{N}_{\mathrm{3D}}^{(\ell)}(i)=
    \left\{j\in\mathcal{C}^{(\ell-1)}:d_{\mathrm{sp}}(j,i)\le r_{\mathrm{nms}}\right\},
    \label{eq:inter_neighborhood}
\end{equation}
where $d_{\mathrm{sp}}$ is the Chebyshev distance in the volume's physical coordinate system with each axis measured in units of its effective voxel pitch, so that a single radius $r_{\mathrm{nms}}$ induces different depth and in-plane extents. Starting with $\mathcal{S}_{\mathrm{inter}}^{(0)}=\varnothing$, each round selects the highest-scoring admissible token and suppresses its neighborhood:
\begin{equation}
    \begin{aligned}
    j_\ell^*&=\arg\max_{j\in\mathcal{C}^{(\ell-1)}}s_j,\\
    \mathcal{S}_{\mathrm{inter}}^{(\ell)}
    &=\mathcal{S}_{\mathrm{inter}}^{(\ell-1)}\cup\{j_\ell^*\},\\
    \mathcal{C}^{(\ell)}
    &=\mathcal{C}^{(\ell-1)}\setminus\mathcal{N}_{\mathrm{3D}}^{(\ell)}(j_\ell^*).
    \end{aligned}
    \label{eq:inter_nms}
\end{equation}
The score $s_j$ remains fixed while the admissible set changes. The procedure stops at $K_{\mathrm{inter}}$ tokens or an empty candidate set. Its output $\mathcal{S}_{\mathrm{inter}}$ thus favors strong but spatially dispersed deviations; these retrieved tokens retain their coordinates and bypass folding.

\subsection{Window-Local Token Folding}
\label{sec:merge}

The selected anchors also provide a structured destination for information that would otherwise be discarded. After excluding $\mathcal{S}_{\mathrm{intra}}$ and $\mathcal{S}_{\mathrm{inter}}$, each remaining token is assigned to the closest intra-slice anchor of its own adaptive window. This restriction avoids folding tokens across the depth discontinuities identified in Eq.~\ref{eq:window_boundary}. Tokens whose closest anchor is too far away are dropped, while retrieved tokens are excluded from folding and remain independent.

For an intra-slice anchor $c$, the folding set $\mathcal{O}(c)$ is then the tokens whose closest anchor is $c$ and whose in-plane Manhattan distance to $c$ is at most a neighborhood radius $\kappa$, and CARVE computes
\begin{equation}
    \begin{aligned} 
        \tilde{\mathbf{f}}_c
        &= \sum_{b \in \mathcal{O}(c)} \alpha_{bc} \mathbf{f}_b, \\
        \alpha_{bc}
        &= \frac{\exp(\cos(\mathbf{f}_b,\mathbf{f}_c)/\tau_m)}
        {\sum_{b'\in\mathcal{O}(c)}\exp(\cos(\mathbf{f}_{b'},\mathbf{f}_c)/\tau_m)}.
    \end{aligned}
    \label{eq:merge}
\end{equation}
where $\tau_m$ is the folding temperature. CARVE finally combines the updated anchor tokens and retrieved tokens as one compressed visual sequence, preserving their respective 3D position indices and the backbone's spatial order. The sequence is projected by $P$, concatenated with tokenized text embeddings, and passed to the LLM.

\begin{table}[!t]
\centering
\scriptsize  
\setlength{\tabcolsep}{1.8pt}  
\renewcommand{\arraystretch}{0.9}

\begin{tabular*}{\columnwidth}{@{\extracolsep{\fill}}lcccccccc@{}}
\toprule
& \multicolumn{5}{c}{Quality} & \multicolumn{3}{c}{Efficiency}\\
\cmidrule(lr){2-6}\cmidrule(lr){7-9}
Method & R-1 & R-L & MTR & BERT & Rel. & Keep & Lat.$\downarrow$ & $\Delta$Mem$\downarrow$ \\
\midrule
Full & 43.71 & 41.28 & 29.08 & 28.70 & 100.0 & 100.0\% & 9.21 & 3.79  \\
\midrule
VisionZip & {36.31} & 34.25 & 22.43 & 20.01 & 78.22 & 20.0\% & 6.90 & 2.84  \\
DivPrune & 36.79 & 34.70 & \underline{23.13} & 21.75 & \underline{80.89} & 20.0\% & 7.22 & \textbf{1.35} \\
FastVID & 31.75 & 30.05 & 19.56 & 15.60 & 66.76 & 20.0\% & 6.70 & 2.84  \\
MMTok & 34.65 & 32.87 & 21.63 & 19.59 & 75.38 & 20.0\% & 13.04 & \underline{1.63}  \\
MedPruner & \underline{36.89} & \underline{35.08} & 21.80 & \underline{21.94} & 80.20 & 21.4\% & \textbf{5.88} & 2.93 \\
\textbf{CARVE} & \textbf{39.04} & \textbf{36.84} & \textbf{24.95} & \textbf{24.09} & \textbf{87.07} & 19.3\% & \underline{6.10} & 2.89  \\
\bottomrule
\end{tabular*}
\caption{AMOS-MM report generation on Hulu-Med-7B at target $r{=}0.2$. Rel.\ is mean retention over the four quality metrics relative to Full; Keep is the realized keep ratio~(\%), while latency and $\Delta$Mem use seconds and GB.}
\label{tab:amos_report_main}
\end{table}
 
\begin{figure}[!t]
\centering
\includegraphics[width=\columnwidth]{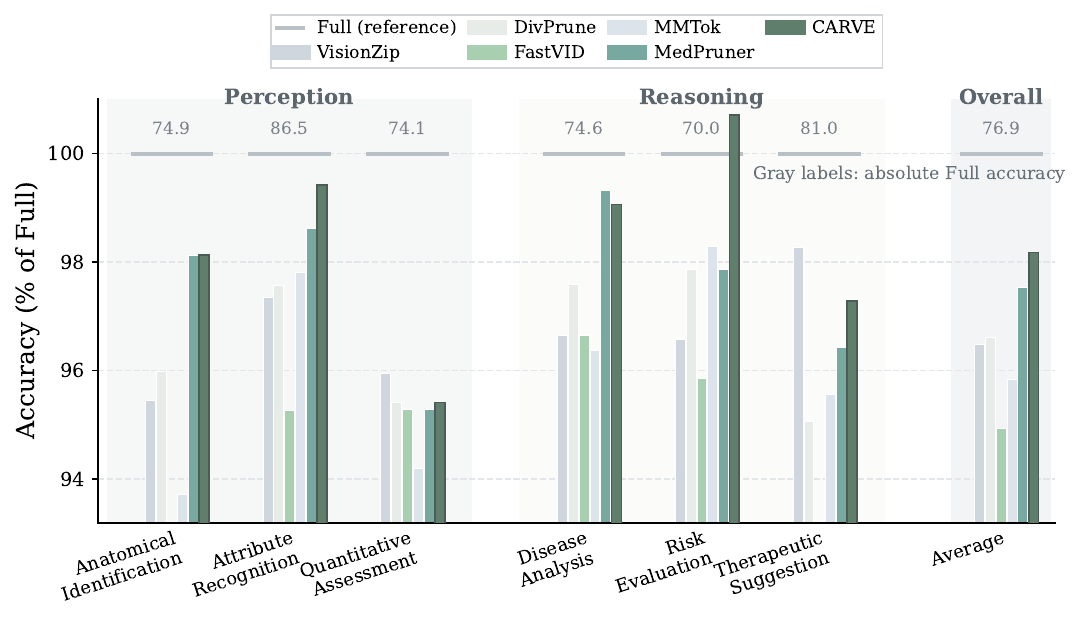}
\caption{AMOS-MM closed-ended VQA by task type on Hulu-Med-7B at target $r{=}0.2$. Bars are relative to Full; gray labels give absolute Full accuracy.}
\label{fig:amos_taskwise}
\end{figure}

% ============================================================
\section{Experiments}
\label{sec:experiments}

\subsection{Experimental Setting}

We evaluate CARVE on three benchmarks spanning closed-ended VQA, open-ended VQA, and report generation: 3D-RAD~\cite{gai20263d}, M3D-VQA~\cite{bai2024m3d}, and AMOS-MM VQA/Report~\cite{ji2022amos}. Unless otherwise stated, the main comparisons use the medical Hulu-Med-7B at a target keep ratio of $r{=}0.2$, which reserves a fraction $\rho{=}0.25$ of the budget for inter-slice retrieval and targets $K_{\mathrm{slice}}{=}64$ anchors per retained slice. The transfer study adds the general-domain Qwen3-VL. We compare against training-free image-token compression methods VisionZip~\cite{yang2025visionzip}, DivPrune~\cite{alvar2025divprune}, and MMTok~\cite{dong2025mmtok}; the video token-compression method FastVID~\cite{shen2026fastvid}; and MedPruner~\cite{liu2026medpruner}, the prior method tailored to medical volumes.

\subsection{Main Results}

\paragraph{AMOS-MM report generation evaluation.}

CARVE ranks first among compressed methods on all four report metrics and their aggregate, while using the smallest realized keep ratio (19.3\%). Report generation separates compression strategies sharply: baseline aggregate retention spans 66.76 to 80.89, and CARVE reaches 87.07---6.18 points above the best baseline aggregate and 2.15 points above the best baseline ROUGE-1. Latency further shows that selector cost is a separate axis: CARVE runs at 6.10\,s, second only to MedPruner and below the 9.21\,s Full reference, whereas MMTok's 13.04\,s exceeds applying no compression at all (Fig.~\ref{fig:tradeoff}).

\paragraph{3D-RAD and M3D-VQA VQA evaluation.}

\begin{table}[t]
\centering
\scriptsize
\setlength{\tabcolsep}{1.5pt}
\renewcommand{\arraystretch}{0.9}

\begin{tabular*}{\columnwidth}{@{\extracolsep{\fill}}lccccccccc@{}}
\toprule
& \multicolumn{6}{c}{Quality} & \multicolumn{3}{c}{Efficiency}\\
\cmidrule(lr){2-7}\cmidrule(lr){8-10}
Method & ACC & B-1 & R-L & MTR & BERT & Rel. & Keep & Lat.$\downarrow$ & $\Delta$Mem$\downarrow$\\
\midrule 
\rowcolor{gray!12}\multicolumn{10}{c}{\textbf{3D-RAD} \textit{(closed + open)}}\\
\midrule
Full        & 79.08 & 25.26 & 33.91 & 20.52 & 55.40 & 100.0 & 100.0\% & 0.62  & 0.97 \\
\midrule
VisionZip   & 78.31 & \underline{23.16} & \underline{32.00} & 19.22 & \underline{53.58} & \underline{95.09} & 20.0\% & 0.36  & \underline{0.40} \\
DivPrune    & 78.00 & 22.88 & 31.48 & 19.26 & 53.53 & 94.51 & 20.0\% & 0.39  & \textbf{0.34} \\
FastVID     & 77.85 & 22.93 & 31.49 & \underline{19.45} & 53.35 & {94.63} & 19.6\% & \underline{0.34}  & \underline{0.40} \\
MMTok       & 77.57 & 22.39 & 30.82 & 18.76 & 53.13 & 92.99 & 20.0\% & 0.43  & \textbf{0.34} \\
MedPruner   & \underline{78.57} & 22.93 & 31.32 & 19.18 & 53.53 & 94.52 & 21.3\% & 0.36 & 0.43 \\
\textbf{CARVE} & \textbf{78.73} & \textbf{23.85} & \textbf{32.56} & \textbf{20.02} & \textbf{53.90} & \textbf{96.97} & 19.7\% & \textbf{0.31} & 0.42 \\
\midrule
\rowcolor{gray!12}\multicolumn{10}{c}{\textbf{M3D-VQA} \textit{(closed + open)}}\\
\midrule
Full        & 77.65 & 45.23 & 47.40 & 31.22 & 56.85 & 100.0 & 100.0\% & 2.36  & 3.23 \\
\midrule
VisionZip   & 76.85 & 44.40 & 46.49 & 30.24 & 56.50 & 98.29 & 20.0\% & 1.10  & 2.31 \\
DivPrune    & 76.79 & 44.56 & 46.68 & 30.56 & 56.50 & 98.63 & 20.0\% & 1.25 & \textbf{1.15} \\
FastVID     & 76.78 & 44.46 & 46.48 & 30.22 & \textbf{56.61} & 98.32 & 20.0\% & \underline{1.08}  & 2.31 \\
MMTok       & \textbf{77.01} & \underline{44.63} & \underline{46.76} & \underline{30.59} & 56.51 & \underline{98.78} & 20.0\% & 7.26  & \underline{1.32} \\
MedPruner   & 76.90 & 44.45 & 46.57 & 30.33 & 56.49 & 98.41 & 19.7\% & \textbf{1.05} & 2.39 \\
\textbf{CARVE} & \underline{76.97} & \textbf{44.83} & \textbf{46.98} & \textbf{30.74} & \underline{56.60} & \textbf{99.08} & 19.2\% & 1.09 & 2.35 \\
\bottomrule
\end{tabular*}
\caption{Main VQA comparison on Hulu-Med-7B at target $r{=}0.2$. Rel.\ is mean retention over the five quality metrics relative to Full; Keep is the realized keep ratio~(\%), while latency and $\Delta$Mem use seconds and GB.}
\label{tab:main}
\end{table}  

At only 19.7\% and 19.2\% realized keep ratios (Table~\ref{tab:main}), CARVE maintains 96.97\% and 99.08\% aggregate performance on 3D-RAD and M3D-VQA, respectively. Closed-ended accuracy is near-saturated at this budget---all compressed methods lie within 1.16 and 0.23 ACC points on the two benchmarks---so the discriminative signal lies in the open-ended metrics, where CARVE ranks first on seven of the eight scores across both benchmarks; the sole exception is M3D-VQA BERT (56.60 vs.\ 56.61). Separation is wider on 3D-RAD, whose $256^2$ inputs leave a more binding absolute budget than the $512^2$ inputs of M3D-VQA at the same ratio.

\paragraph{AMOS-MM VQA task-type evaluation.}

Across three perception and three reasoning task types (Fig.~\ref{fig:amos_taskwise}), CARVE achieves the highest overall relative score among compressed methods and does not regress on either group, indicating that its gains are not confined to coarse perception.

\paragraph{Backbone transfer.}

To test whether CARVE depends on a particular projector--LLM stack, we evaluate three additional backbones on six tracks at the same target keep ratio.
\begin{figure}[t]
\centering
\includegraphics[width=0.9\columnwidth]{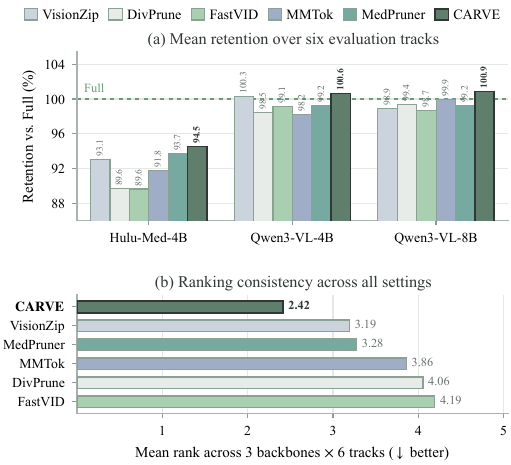}
\caption{Cross-backbone transfer at target $r{=}0.2$. (a) Mean retention over six evaluation tracks relative to Full. (b) Mean rank over all $3$ backbones $\times$ $6$ tracks among compressed methods (lower is better). CARVE achieves the highest aggregate retention on every backbone and the best overall mean rank.}
\label{fig:backbone_transfer}
\end{figure}
\begin{figure}[!t]
\centering 
\includegraphics[width=0.9\columnwidth]{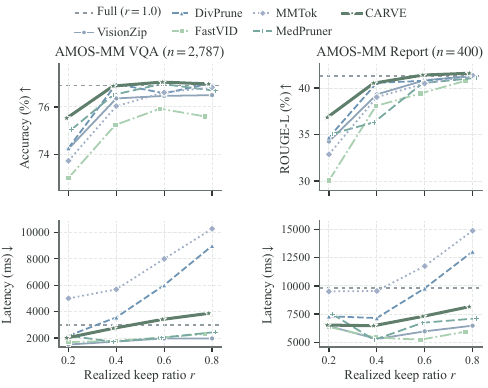} 
\caption{Quality and latency versus realized keep ratio on AMOS-MM VQA and Report. Dashed lines mark the uncompressed Full reference.} 
\label{fig:kr_plot}
\end{figure}
CARVE is first on all three (Fig.~\ref{fig:backbone_transfer}): 94.53\% on Hulu-Med-4B, 100.64\% and 100.88\% on Qwen3-VL-4B/8B, with a mean rank of 2.42 across the 18 backbone--track combinations against 3.19 for the next method. Separation is widest on Hulu-Med-4B, the most binding setting, whereas the Qwen3-VL routes leave every method near the Full reference, indicating that 20\% is not yet a binding budget for that stack.

\subsection{Ablation Studies}
\label{sec:ablation}

We organize the ablation into four comparison groups: naive sampling; depth partition and window allocation; intra/inter selection roles; and token folding. All variants use the same target budget $K$, preprocessing, decoder settings, and evaluation samples on AMOS-MM, and paired rows isolate the intended factor within each group.
\begin{table}[t]
\centering
\scriptsize 
\setlength{\tabcolsep}{1.1pt}
\renewcommand{\arraystretch}{0.9} 
\begin{tabular*}{\columnwidth}{llc@{\extracolsep{\fill}}cccc@{}}
\toprule
 & & & \multicolumn{2}{c}{VQA} & \multicolumn{2}{c}{Report} \\
\cmidrule(lr){4-5}\cmidrule(lr){6-7}
Control & Variant & Keep & ACC & Lat.$\downarrow$ & R-L & Lat.$\downarrow$ \\
\midrule
 & \textbf{CARVE (Ours)} & {19.3\%} & \textbf{75.50} & 1.11 & \textbf{36.84} & 6.10 \\
\midrule 
\multicolumn{7}{@{}l}{\emph{Anchor: Naive sampling}}\\
& Uniform slice+grid & 20.0\% & 73.46$_{\textcolor{gray}{\downarrow\,2.04}}$ & 0.94 & 33.45$_{\textcolor{gray}{\downarrow\,3.39}}$ & 5.75 \\ 
\cmidrule(lr){1-7}
\multicolumn{7}{@{}l}{\emph{A: Depth partition and window allocation}}\\
& Adaptive win., uniform bud. & 19.3\% & 74.45$_{\textcolor{gray}{\downarrow\,1.05}}$ & 1.10 & 35.18$_{\textcolor{gray}{\downarrow\,1.66}}$ & 6.10 \\
& Uniform win., uniform bud. & 19.6\% & 74.42$_{\textcolor{gray}{\downarrow\,1.08}}$ & 1.11 & 34.68$_{\textcolor{gray}{\downarrow\,2.16}}$ & 6.08 \\
\cmidrule(lr){1-7} 
\multicolumn{7}{@{}l}{\emph{B: Intra/inter selection roles}}\\
& Intra only ($\rho=0$) & 20.3\% & 74.49$_{\textcolor{gray}{\downarrow\,1.01}}$ & 1.13 & 35.15$_{\textcolor{gray}{\downarrow\,1.69}}$ & 5.99 \\
& Intra only, w/o quadtree & 19.1\% & 74.24$_{\textcolor{gray}{\downarrow\,1.26}}$ & 1.36 & 34.67$_{\textcolor{gray}{\downarrow\,2.17}}$ & 6.12 \\
& Inter only & 19.3\% & 74.74$_{\textcolor{gray}{\downarrow\,0.76}}$ & 1.13 & 34.92$_{\textcolor{gray}{\downarrow\,1.92}}$ & 6.15 \\
& Inter only, w/o 3D-NMS & 20.1\% & 74.20$_{\textcolor{gray}{\downarrow\,1.30}}$ & 1.14 & 34.22$_{\textcolor{gray}{\downarrow\,2.62}}$ & 5.82 \\
\cmidrule(lr){1-7}
\multicolumn{7}{@{}l}{\emph{C: Token folding}}\\
& Direct drop & 19.3\% & 74.13$_{\textcolor{gray}{\downarrow\,1.37}}$ & 1.29 & 33.44$_{\textcolor{gray}{\downarrow\,3.40}}$ & 5.87 \\
& Global merge & 19.3\% & 74.20$_{\textcolor{gray}{\downarrow\,1.30}}$ & 1.38 & 35.33$_{\textcolor{gray}{\downarrow\,1.51}}$ & 6.30 \\
& Local mean merge & 19.3\% & 75.06$_{\textcolor{gray}{\downarrow\,0.44}}$ & 1.11 & 35.26$_{\textcolor{gray}{\downarrow\,1.58}}$ & 6.03 \\
\bottomrule 
\end{tabular*}
\caption{AMOS-MM ablation at target $r{=}0.2$. Keep is the realized ratio~(\%); the remaining metrics are ACC, R-L, and latency~(s). Subscripts give the drop relative to the full method, which is set in bold as the reference; all other rows are ablated versions of it, so no ranking among variants is implied.}
\label{tab:ablation}
\end{table} 

Every toggle costs quality on both tasks (Table~\ref{tab:ablation}). Naive Sampling has the largest ACC drop, while Dropping Folding has the largest ROUGE-L drop. Group A separates the allocation factors: adaptive windows with a uniform budget cost 1.66 ROUGE-L, and uniform windows a further 0.50, so the allocation matters more than the partition alone. In group B, neither branch suffices by itself (1.69 and 1.92 ROUGE-L).

\subsection{Ratio Change and Efficiency Analysis}

Figure~\ref{fig:kr_plot} sweeps the target ratio over $r\in\{0.2,0.4,0.6,0.8\}$ and plots quality and latency against the realized keep ratio. Quality rises with the budget for every method, but the methods converge: the spread narrows from 2.51 to 1.40 ACC on VQA and from 6.80 to 0.82 ROUGE-L on report generation. CARVE's advantage is concentrated where the budget binds---it leads by 0.48 ACC and 1.76 ROUGE-L at $r{\approx}0.2$ and stays within 0.1 of the best baseline above that---while both tracks return to near-Full quality by $r{\approx}0.4$. Latency exposes the cost side: CARVE remains below the Full reference on report generation at every ratio, whereas MMTok exceeds it throughout and DivPrune from $r{\approx}0.6$.

\subsection{Qualitative Analysis}  

\begin{figure}[!t]
\centering  
\includegraphics[width=\columnwidth]{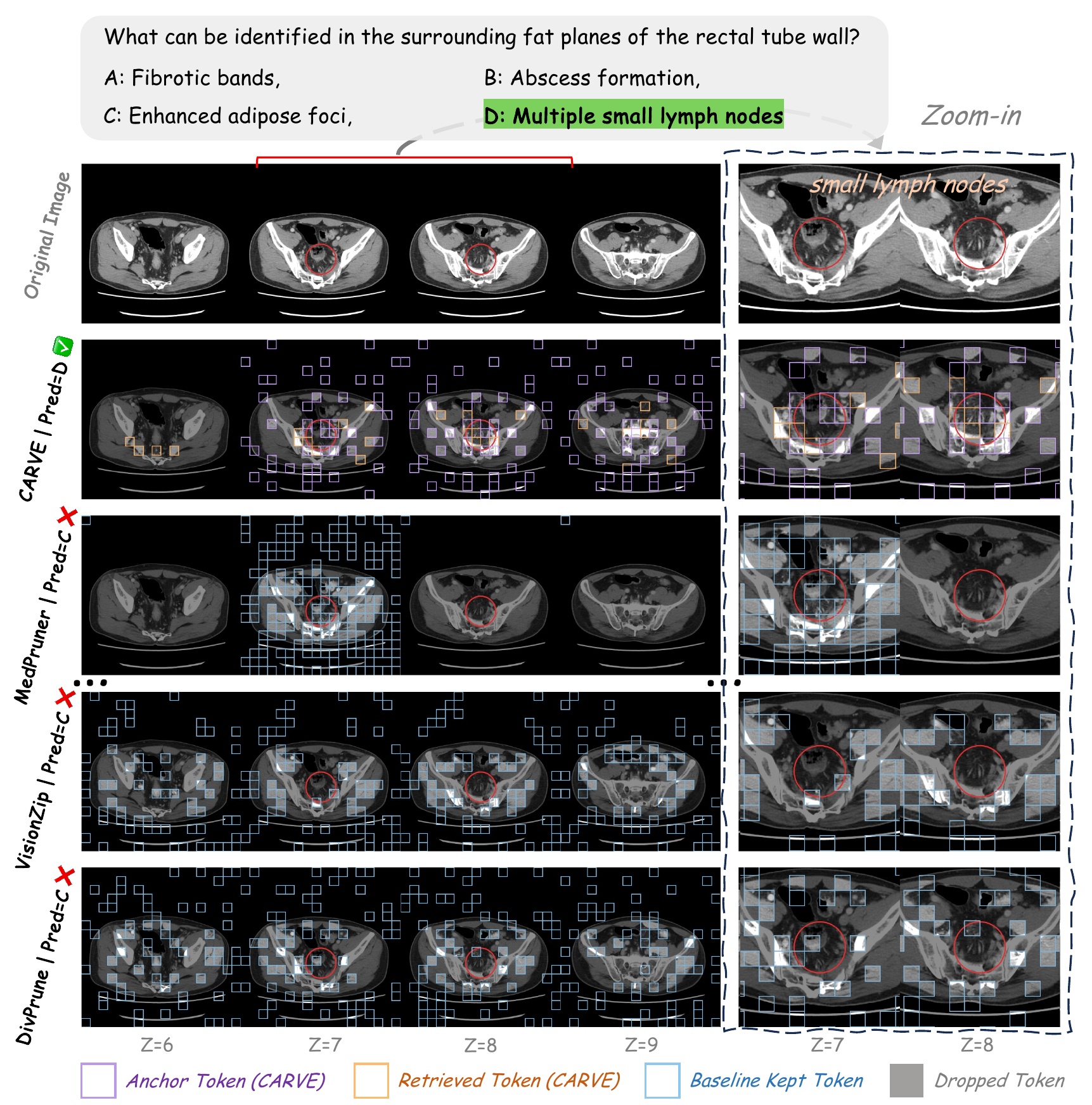}
\caption{Retained tokens on four consecutive slices of one AMOS-MM case, with two key slices magnified. Purple and orange mark CARVE's anchor and retrieved tokens, blue marks baseline kept tokens, and red circles indicate the queried finding.}
\label{fig:vis}  
\end{figure}
 
Figure~\ref{fig:vis} shows a case whose answer depends on a small finding that appears on adjacent slices. CARVE spreads anchors across all four slices, and in the two magnified planes its retrieved tokens concentrate on the queried lymph nodes, whose appearance fluctuates from one slice to the next; it is the only method that answers correctly. MedPruner spends almost its entire budget on a single slice and leaves the neighboring ones empty, following from its whole-slice granularity, whereas VisionZip and DivPrune distribute tokens evenly and cover the finding only incidentally.
 
% ============================================================ 
\section{Conclusion}
\label{sec:conclusion}

We presented CARVE, a training-free token reallocation framework for slice-based 3D medical vision-language models that spends one budget anisotropically across the depth and in-plane axes before the frozen projector and LLM. At approximately 20\% retained tokens, it leads all compressed methods on the AMOS-MM report metrics and attains the highest aggregate retention across backbones, without training or backbone modification. Future work may combine CARVE with intra-ViT token merging or learnable budget allocation, extend it to multimodal imaging such as PET-CT.
 
% ============================================================
\bibliography{references}

\end{document}